\documentclass[10pt,twocolumn,letterpaper]{article}

\usepackage{iccv}              
\usepackage{algpseudocode}
\usepackage[accsupp]{axessibility}
\usepackage{algorithm}
\usepackage{stfloats}
\usepackage{cuted}
\usepackage{multirow}
\usepackage{color}

\usepackage{multirow}
\usepackage{array}
\usepackage{algorithm}
\usepackage{algpseudocode}
\usepackage[shortlabels]{enumitem}
\usepackage{stfloats}
\newcommand{\myparagraph}[1]{\vspace{0.1cm}\noindent\textbf{#1}}

\newcommand{\corrauthor}{\textsuperscript{\textdagger}}

\definecolor{iccvblue}{rgb}{0.21,0.49,0.74}
\usepackage[pagebackref,breaklinks,colorlinks,allcolors=iccvblue]{hyperref}
\definecolor{r1}{RGB}{230, 120, 40}
\definecolor{r2}{RGB}{50, 90, 180}
\definecolor{r3}{RGB}{0, 140, 130}
\def\paperID{7965} 
\def\confName{ICCV}
\def\confYear{2025}

\title{ScenePainter: Semantically Consistent Perpetual 3D Scene \protect\\ Generation with Concept Relation Alignment}

\begin{document}
\vspace{-3mm}
\author{
Chong Xia,
Shengjun Zhang,
Fangfu Liu,
Chang Liu, \\
Khodchaphun Hirunyaratsameewong,
Yueqi Duan\corrauthor\\
Tsinghua University
}



\maketitle

\begingroup
\renewcommand\thefootnote{} 
\footnote{\corrauthor Corresponding author.}
\endgroup


\begin{strip}
\vspace{-20mm}
    \centering
    \includegraphics[width=1.0\linewidth]{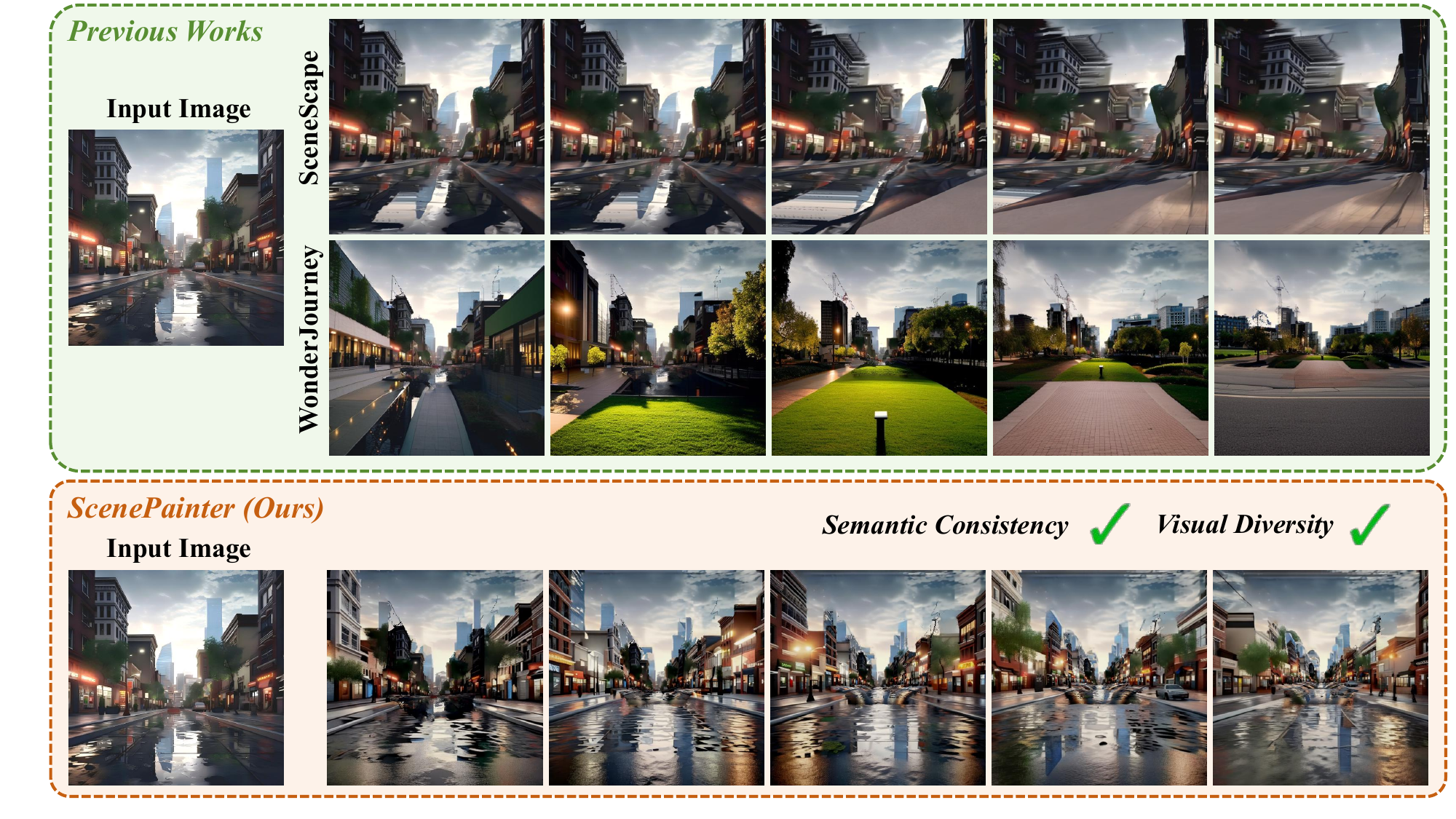}
    \captionof{figure}{ \textbf{We propose ScenePainter, which aims to generate semantically consistent yet visually diverse 3D view sequences starting from a single view.} We show painted images from the view of a moist and shadowed street with a receding 3D camera motion. ScenePainter can generate plausible and consistent content given the first view while maintaining diversity in forms and appearances.}
    \label{teaser}
\end{strip}

\begin{abstract}

    Perpetual 3D scene generation aims to produce long-range and coherent 3D view sequences, which is applicable for long-term video synthesis and 3D scene reconstruction. Existing methods follow a ``navigate-and-imagine" fashion and rely on outpainting for successive view expansion. However, the generated view sequences suffer from semantic drift issue derived from the accumulated deviation of the outpainting module. To tackle this challenge, we propose ScenePainter, a new framework for semantically consistent 3D scene generation, which aligns the outpainter's scene-specific prior with the comprehension of the current scene. To be specific, we introduce a hierarchical graph structure dubbed SceneConceptGraph to construct relations among multi-level scene concepts, which directs the outpainter for consistent novel views and can be dynamically refined to enhance diversity. Extensive experiments demonstrate that our framework overcomes the semantic drift issue and generates more consistent and immersive 3D view sequences. Project Page: \url{https://xiac20.github.io/ScenePainter/}.

\end{abstract}

\section{Introduction}
With the rapid development of diffusion models, 3D generation has become one of the most significant problems in 3D computer vision. One main branch perpetual 3D scene generation has garnered more attention from academia and industry for their potential in large 3D scene construction and long video synthesis. Given an initial single image or a scene description, perpetual 3D scene generation aims to generate a series of 3D scene views, which are consistent in geometry and semantic relations. To achieve this, previous works mainly follow a 'navigate and imagine' paradigm, which can be decomposed into the iterative processes of unprojecting, rendering, and outpainting. The geometry consistency problem has been well solved by previous works due to the improved monocular depth estimation and effective 3D scene representations like point cloud, mesh or 3D gaussian splatting~\cite{kerbl20233d}. 

However, the semantic consistency problem has not received enough attention, and causes serious semantic drift issue derived from the accumulated deviation and artifacts of the native outpainting process. One of previous state-of-the-art models WonderJourney~\cite{yu2024wonderjourney} frequently encounters semantic drift issue, shifting from the arid desert to the vibrant lake or shifting from lush green scenes to snow-covered landscapes. Although this work is dedicated to generating diverse views, the scene inconsistency problem remains particularly pronounced and affects the scene quality. In addition, the semantic drift issue is bypassed in earlier works like  InfiniteNature-Zero~\cite{li2022infinitenature} and SceneScape~\cite{fridman2024scenescape}, as they maintain subtle camera movements between adjacent frames for smooth geometric transformation, which hinders the generation of novel objects and leads to monotonous scenes. Therefore, generating semantically consistent and content-rich scene views is a key challenge in 3D scene generation as shown in Figure \ref{teaser}.

\begin{figure*}[t]
    \centering
    \includegraphics[width=\linewidth]{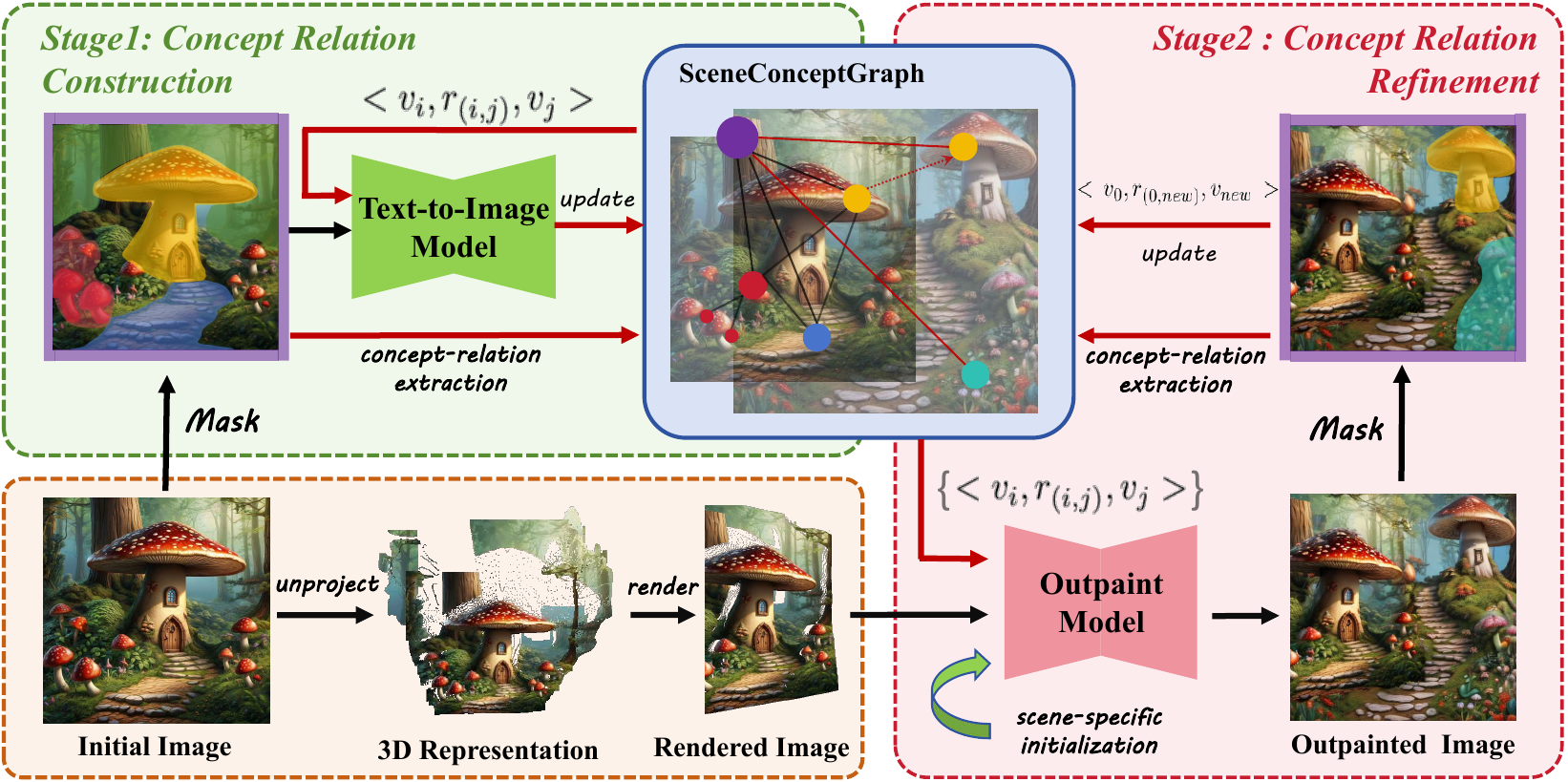}
    \caption{\textbf{Overall framework of our approach ScenePainter}.  We propose a two-stage framework that constructs and refines scene concept relations with the graph structure SceneConceptGraph, and aligns the outpainting model with the scene-specific prior during the ongoing painting process. We use colored dots and regions to represent scene concept nodes and their masks, and black and red lines in SceneConceptGraph to denote the initial and newly added relation edges. Here, we simplify the update process in the second stage, which is similar to the one in the first stage.}
    \label{fig:pipeline}
\end{figure*}

Towards this goal, we propose ScenePainter, a two-stage framework for semantically consistent yet visually diverse 3D scene generation illustrated in Figure \ref{fig:pipeline}. In the first stage of concept relation construction, we propose a scene-level customization method to achieve a thorough comprehension of the current scene. Unlike previous single image or multi-concept customization methods, scenes contain richer concepts and more complicated relations among them, such as the relative spatial position of objects, semantic distribution, spatial layout and overall style, which makes achieving satisfactory customization results particularly challenging. To solve this problem, we first extract multilevel scene concepts rather than simply multiple objects, and then leverage a hierarchical graph structure dubbed SceneConceptGraph to further construct the relations among concepts. For each concept-relation pair in the SceneConceptGraph, we take it as dedicated textual embeddings for text-to-image models to restore the original image region referenced by the relation. In this way, we obtain scene-specific textual embeddings of each concept-relation pair and optimized text-to-image model, which jointly direct the subsequent outpainting model with scene-specific prior transmission.

In the second stage of concept relation refinement, we follow the mainstream 3D generation framework and align the native outpainting model with extracted scene-specific prior to achieve semantically consistency and mitigate generation deivation. To be specific, we initialize the outpainter with the aforementioned optimized text-to-image model and
guide the outpainting process with updated concept-relation pair. Further, generating nearly identical scenes would lead to repetition and monotony, thus we strive to enhance visual diversity while ensuring semantic consistency. During the outpainting process, we can dynamically add new objects, change existing objects, or smoothly transition to other scenes with test-time refinement of the SceneConceptGraph. In this way, users can easily enrich and expand the current scene while enabling novel creation, thereby achieving both consistency and diversity simultaneously. Our main contributions are as follows:
\vspace{0.1cm}
\begin{itemize}
    \item We design a scene-level customization method, which constructs the concepts and relations of the scene and generates consistent scene views.
    \vspace{0.05cm}
    \item We propose a new 3D scene generation framework ScenePainter, which aligns the outpainting model with scene-specific prior for consistent and diverse scene expansion.  
    \item Extensive experiments demonstrate that our framework produces more consistent and vivid 3D view sequences in comparison to previous state-of-the-art works.
    
\end{itemize}

\section{Related Work}

\myparagraph{3D Scene Generation.} Due to the recent success of image diffusion models~\cite{ho2020denoising, rombach2022high, podell2023sdxl}, 3D scene generation has seen rapid developments. Existing methods can be mainly divided into two branches. One branch of methods~\cite{wang2024customizing, zhang2024taming, zhou2025dreamscene360, zhou2024holodreamer, yang2024layerpano3d} employ 2D panorama to represent panoramic environments and lift it to 3D space. However, limited by the representative capacity of 2D panoramas, these methods often result in blurry renderings, ambiguity and gaps in 3D scene. Moreover, they fail to generate arbitrary 3D scene following specific camera trajectory or interactive user instructions, which limits their applications.

The other branch of methods ~\cite{kaneva2010infinite, chai2023persistent, cai2023diffdreamer} seeks to create perpetual scenes by leveraging a ``navigate-and-imagine" paradigm, which renders the partial image, outpaints for a complete next view and repeats the process. The foundational works, Infinite Images~\cite{kaneva2010infinite}, Infinite Nature~\cite{liu2021infinite} and InfiniteNature-Zero~\cite{li2022infinitenature}, first propose the autoregressive pipeline. SceneScape~\cite{fridman2024scenescape} further exploits the generative power and incrementally constructs a single cave-like environment. A recent development, WonderJourney~\cite{yu2024wonderjourney}, focuses on creating diverse view sequences combining off-the-shelf models. While these approaches explore scene generation, they either generate diverse scene with obvious semantic drift issue or generate scenes with limited content and viewpoints. In contrast, our approach generates scene-consistent and content-rich view sequences, achieving the balance between consistency and diversity.

\myparagraph{Customized Generation.} Customized image generation has attracted increasing attention due to its ability to create consistent images with user preferences. Two fundamental methods in text-to-image model customization are Textual Inversion (TI)~\cite{gal2022image} and DreamBooth (DB)~\cite{ruiz2023dreambooth}, with TI focusing on optimizing text embeddings, while DB fine-tuning the whole image diffusion model. Other notable approaches include Custom Diffusion~\cite{kumari2023multi}, which fine-tunes only specific layers, and LoRA~\cite{hu2021lora, ryulow}, which restricts updates to rank 1 matrices. Expanding on these techniques, various studies ~\cite{voynov2023p+,wei2023elite,han2023svdiff, ye2023ip, shi2024instantbooth} have explored single subject or multi-object customization from a single or several reference images. Break-A-Scene~\cite{avrahami2023break} extracts multiple concepts from one single image and employs masks to indicate different objects. To the best of our knowledge, we are the first work to achieve scene-level customization by constructing scene graph and utilizing concept-relation pair textual embeddings, thus creating consistent scene views.

\myparagraph{Video Generation.} 
Expanded upon basic text-to-image diffusion models~\cite{ho2020denoising, rombach2022high, podell2023sdxl}, recent works on text-to-video and image-to-video generation~\cite{chen2023videocrafter1, brooks2024video, blattmann2023stable} have seen rapid advancements by incorporating temporal dynamics, enabling the production of high-quality and diverse video content. However, many of these approaches~\cite{wang2024motionctrl, jain2024peekaboo, yatim2024space} focus on the dynamics of the subject with a relatively static viewpoint, such as human walking or pets running. Even with delicate text prompt, these methods are reluctant to create more scene content with large viewpoint transition. In this work, we mainly focus on long-range 3D view sequences generation with free-moving camera trajectories and rich content, which may potentially serve as keyframes for synthesizing long-term 3D videos.


\section{Approach}
In this section, we present our method ScenePainter for 3D scene generation. At first, we review the general pipeline of perpetual 3D scene generation and briefly illustrate the framework of our ScenePainter in Section \ref{sec:overview}. Next, we provide a detailed introduction to our two main stages: Concept Relation Construction in Section \ref{sec:construction} and Concept Relation Refinement in Section \ref{sec:refinement}.

\subsection{Overview of ScenePainter}
\label{sec:overview}
Perpetual 3D scene generation aims to synthesize a series of consistent yet diverse 3D views, starting from an arbitrary single image and following a long-range specified camera trajectory. The general pipeline mainly consists of three iterative processes: $unproject$, $render$, and $outpaint$, which could be further explained as lifting current 2D image to 3D representation, rendering partial image at next view camera and outpainting for a complete next image. The entire process could be modular, leveraging pre-trained monocular depth estimator, outpainting model and optional vision language model. To mathematically describe this, we suppose the generated view stream denoted as $\mathcal{I}=\{I_1,I_2,...,I_T\}$ with each component $I_i$ derived from $I_{i-1}$ as:

\begin{equation}
\begin{aligned}
    D_{i-1} &= \mathrm{DepthEstimator}(I_{i-1}),  \\
    P_{i-1} &= unproject(I_{i-1},D_{i-1}),  \\
    S_{i}&=S_{i-1}\cup P_{i-1},\\
    \bar{I}_{i},  M_{i}&= render(S_i,C_i),  \\
    I_{i}&=outpaint(\bar{I}_{i},  M_{i}, T_{i}).
\end{aligned}
\end{equation}

Here $D_i$, $C_i$, $\bar{I}_i$, $M_i$, $T_i$ represent the estimated depth, specified camera trajectory, partial image and mask for outpainting, and optional text prompt at $i^{th}$ frame respectively. To ensure geometry consistency, previous works focus on the representation of single 2D image in 3D space and the unified 3D scene representation to establish appropriate geometric structures, corresponding to $P_i$ and $S_i$ as mentioned above. These methods mainly contribute on specific depth refinement strategies and effective 3D representations such as point cloud~\cite{yu2024wonderjourney}, mesh~\cite{fridman2024scenescape} and gaussian surfels~\cite{chung2023luciddreamer, yu2024wonderworld}.

However, beyond geometric consistency, semantic consistency is also a crucial and challenging problem in 3D scene generation. Merely depending on the partial image and optional text prompt, it is difficult for the off-the-shelf outpainting model to provide semantically consistent painted results that accurately match with the existing partial scene, which could be attributed to the limited performance and inherent diversity of the outpainter. Moreover, even though there may be slight deviation in semantic comprehension each time, the semantic drift error will be accumulated and amplified with iterative processing, leading to a significant difference from the original scene definition. This is referred to as the semantic transfer issue. When the first and last images are placed together, they appear to be two views from totally different and separate scenes. 

In order to solve this issue, we propose ScenePainter, which constructs unified and comprehensive understanding of the scene concept relations based on the initial scene definition and aligns the scene-specific prior of the outpainter with it to ensure semantic consistency. During the subsequent view generation process, we continuously refine the understanding of scene relationships while ensuring that the outpainter is updated in parallel, allowing the scene to evolve and enrich in a controllable way. The whole procedure could be depicted as follows:

\begin{equation}
\begin{aligned}
    \bar{\mathcal{M}}_0, G_0 = construct(I_0), \\
    \mathcal{M}_0 = BLD(\bar{\mathcal{M}_0}), \\
\end{aligned}
\end{equation}

\begin{equation}
\begin{aligned}
    I_{i+1} = \mathcal{M}_i(I_i, T_i, G_i), \\
    G_{i+1}, \mathcal{M}_{i+1} = refine(I_{i+1}, G_i, \mathcal{M}_i),\\
\end{aligned}
\end{equation}
where $\bar{\mathcal{M}_0}$ and $\mathcal{M_0}$ denote the initial optimized text-to-image generation model and the converted outpainting model by Blended Latent Diffusion (BLD)~\cite{avrahami2023blended} for scene expansion. $G$ refers to the specific multi-layer graph structure for concept relations extraction dubbed as SceneConceptGraph, as illustrated in Figure \ref{fig:pipeline} . In the following parts, we will explain the construction and refinement process of scene concept relations in detail.

\subsection{Concept Relation Construction}
\label{sec:construction}
Given an initial scene view $I_0$, we construct a relation graph among multi-level scene concepts, $\mathcal{G}=<\mathcal{V}, \mathcal{R}>$, where $\mathcal{V}$ and $\mathcal{R}$ represent the sets of concept vertices and relation edges respectively. The $\mathcal{V}$ contains three-layer concept node sets, \ie $\mathcal{V}=\{\mathcal{V}_1, \mathcal{V}_2, \mathcal{V}_3\}$. The first-level node set $\mathcal{V}_1$ contains only a single concept, which represents the overall environment and style, typically denoted as $v_0$. The second-level node set $\mathcal{V}_2$ encompasses all the areas with the same category which we wish to focus on, such as a forest or a group of buildings. The third-level node set $\mathcal{V}_3$ represents the most fundamental objects in the scene. Specifically, for objects with uniqueness, we assume that they belong to second-level node set $\mathcal{V}_2$, where a single object represents its corresponding category region. And for objects with diverse appearances but belong to the same type, we categorize them as third-level nodes, and the common region formed by multiple objects is assigned to the second-level node. Besides, for each concept, we store its corresponding region mask and for each relation edge, we consider the combined area of the two connected concepts as its mask. We maintain the region affiliation mapping function, so that for any third-level node, the corresponding second-level node can be identified, which can be formulated as:
\begin{equation}
    \forall v_i \in \mathcal{V}_3,\  \exists v_j \in \mathcal{V}_2,\  \text{s.t.}\  v_j = region(v_i)
\end{equation}

The relation edge set $\mathcal{R}$ consists of three types of relations: $\mathcal{R}_1$ connects vertices in $\mathcal{V}_1$ and $\mathcal{V}_2$ and represents the spatial layout and style of category regions in the whole scene; $\mathcal{R}_2$ connects vertices both in $\mathcal{V}_2$ and represents the relative spatial layout and semantic connection of different category regions; $\mathcal{R}_3$ connects node in $\mathcal{V}_3$ and corresponding region node in $\mathcal{V}_2$ represents the relative spatial layout within the category region. 

Considering the fact that concepts and relations in the scene have much more complicated and comprehensive meanings than plain text description, we decide to adopt learning-based method to optimize the concept and relation embeddings, which is widely used in customization generation. To be specific, we aim to extract $N+M$ textual handles $\{v_i\}_{i=1}^{N}$ and $\{r_i\}_{i=1}^{M}$ from off-the-shelf text-to-image models, s.t. the textual handle represents the corresponding concept or relation. And the optimized handles can then be correspondingly considered as text prompts to guide the synthesis of new scene views and novel combinations of scene concepts and relations. Unlike previous object-centric customization that utilizes basic object handles as text prompt, we focus on relation among concepts and take each relation-concept pair as text guidance, which could be denoted as $<v_i, r_{(i,j)}, v_j>$.

In terms of specific customization strategies, two classic methods are: Textual Inversion (TI)~\cite{gal2022image}, which extracts text embeddings but fails to preserve their identity, and DreamBooth (DB)~\cite{ruiz2023dreambooth}, which fine-tunes the entire diffusion model but lacks diversity. As a result, we optimize both the textual handles and model weights, combining the basic two customization strategies in two different phases as~\cite{avrahami2023break}. The combination enables the textual handles and the diffusion model to collaborate in iterative updates and construct a personalized understanding. The optimized text-to-image generation model with scene-specific prior would serve as the initialization parameters of the outpainter in the following view generation process by the strategy Blended Latent Diffusion~\cite{avrahami2023blended}. 

The handles are optimized using a combination of a masked reconstruction diffusion loss, a scene-specific prior preservation loss, and a cross-attention loss. For each relation-concept handle pair $<v_i,r_{(i,j)},v_j>$ with masks $m_i$ and $m_j$, we keep the two concepts unchanged and outpaint the remaining part to generate novel views with similar scene environment to serve as training samples for scene-specific prior preservation and scene diversity. Then we supervise the reconstruction quality within the union region of the two concepts and the cross-attention map between each handle and its corresponding mask to get the handle focus on the regions it refers to. The total loss formulation can be summarized as:
\begin{equation}
\begin{aligned}
    \mathcal{L}_{\mathrm{rec}} &= \mathbb{E}_{z, s, \epsilon \sim \mathcal{N}(0, 1), t }\Big[ \Vert \epsilon \odot m_u - \epsilon_\theta(z_{t},t,p_s) \odot m_u \Vert_{2}^{2}\Big],\\
        \mathcal{L}_{\mathrm{prior}} &= \mathbb{E}_{z, s, \epsilon \sim \mathcal{N}(0, 1), t }\Big[ \Vert \epsilon - \epsilon_\theta(z_{t},t,p_s) \Vert_{2}^{2}\Big],\\
    \mathcal{L}_{\mathrm{attn}} &= \mathbb{E}_{z, h, t}\Big[ \Vert CA_{\theta}(h, z_t) - m_h \Vert_{2}^{2}\Big],\\
    \mathcal{L}_{\mathrm{total}} &= \mathcal{L}_{\mathrm{rec}} + \lambda_{\mathrm{prior}} \mathcal{L}_{\mathrm{prior}}+     
    \lambda_{\mathrm{attn}} \mathcal{L}_{\mathrm{attn}},
\label{loss}
\end{aligned}
\end{equation}
where $z_t$ is the noisy latent at time step $t$, $p_s$ is the text prompt, $m_u$ is the union of the handle masks $m_h$, $\epsilon$ is the added noise, $\epsilon_\theta$ is the denoising network, and $CA_{\theta}(h, z_t)$ is the cross-attention map between the handle $h$ and the noisy latent $z_t$. $\lambda_{\mathrm{prior}}$ and $\lambda_{\mathrm{attn}}$ are hyperparameters to balance the combination of the total loss function.

After the first stage, we construct the SceneConceptGraph which represents concepts and relations with dedicated textual embeddings and at the same time, we prepare the initialization weights for the outpainter with customized scene-specific prior.

\subsection{Concept Relation Refinement}
\label{sec:refinement}

During the view sequence generation stage, relying on the graph construction of relations among scene concepts, the outpainting module tends to generate novel views consistent with the initial scene definition, which share a unified overall style, similar object characteristics and harmonized spatial layout. Moreover, in terms of scene diversity and editability, unlike previous works that adopt predefined or automatically generated text prompt as outpainting guidance and suffer from the limited prompt fidelity of the outpainter model leading to semantic drift issue, our model are more friendly to user-specified text prompt, with simply describing a handle or several handles to change the appearance or spatial location, muting a handle to keep it from appearing in the next frame and detailed description about new objects to generate novel concepts. 

Furthermore, in order to dynamically refine the SceneConceptGraph with new concept relations and adaptively align the scene understanding of the outpainter for more consistent subsequent views,  we take a test-time training approach to simultaneously update new text embeddings and align the outpainter. Unlike the construction process that focuses on all three types of concepts and relations, we only concentrate on one relation edge between the first-level concept \ie the overall environment and second-level concept specified by users in the refinement stage for efficiency and real-time capability. To be specific, for the addition of a new concept, we assign new concept holder $v_n$ and relation holder $r_{(0,n)}$, and then take the relation-concept pair $<v_0, r_{(0,n)}, v_n>$ as textual embeddings for training. And for the change of an existing concept $v_i$, we take the relation-concept pair $<v_0, r_{(0,i)}, v_i>$ accordingly. We use segmentation models to get the assigned concept mask and optimize both the textual handles and model weights with masked diffusion loss and cross-attention loss as mentioned before. The generated view sequences tend to maintain consistent relationships while making versatile changes.
\section{Experiments}

\subsection{Comparison Methods}
 \begin{figure*}[t]
    \centering
    \includegraphics[width=1.0\linewidth]{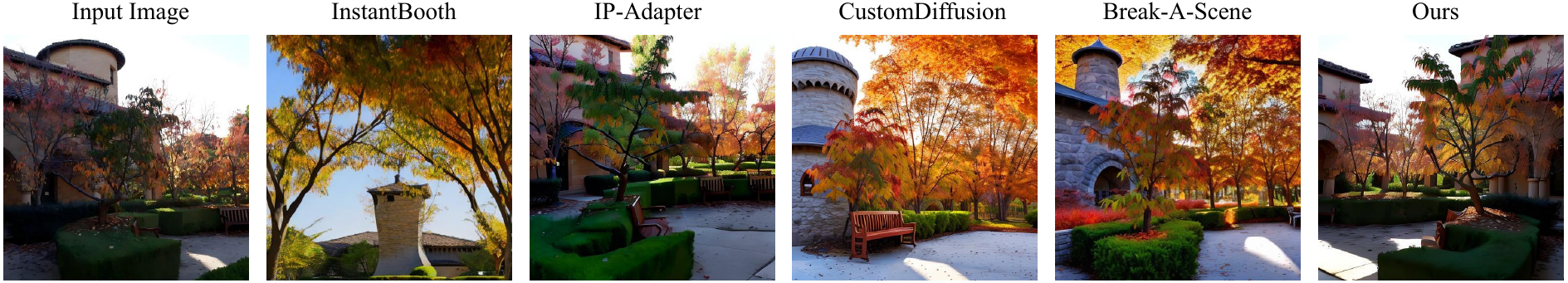}
    \caption{\textbf{Comparison with other customization methods.} InstantBooth and IP-Adapter are designed for the whole image customization without concept mask, while Custom Diffusion and Break-A-Scene focus on multi-concept customization. All these methods fail to achieve satisfactory scene-level customization.}
    \label{fig:comparisonbas}
\end{figure*}

 \begin{figure*}[t]
    \centering
    \vspace{-3mm}
    \includegraphics[width=1.0\linewidth]{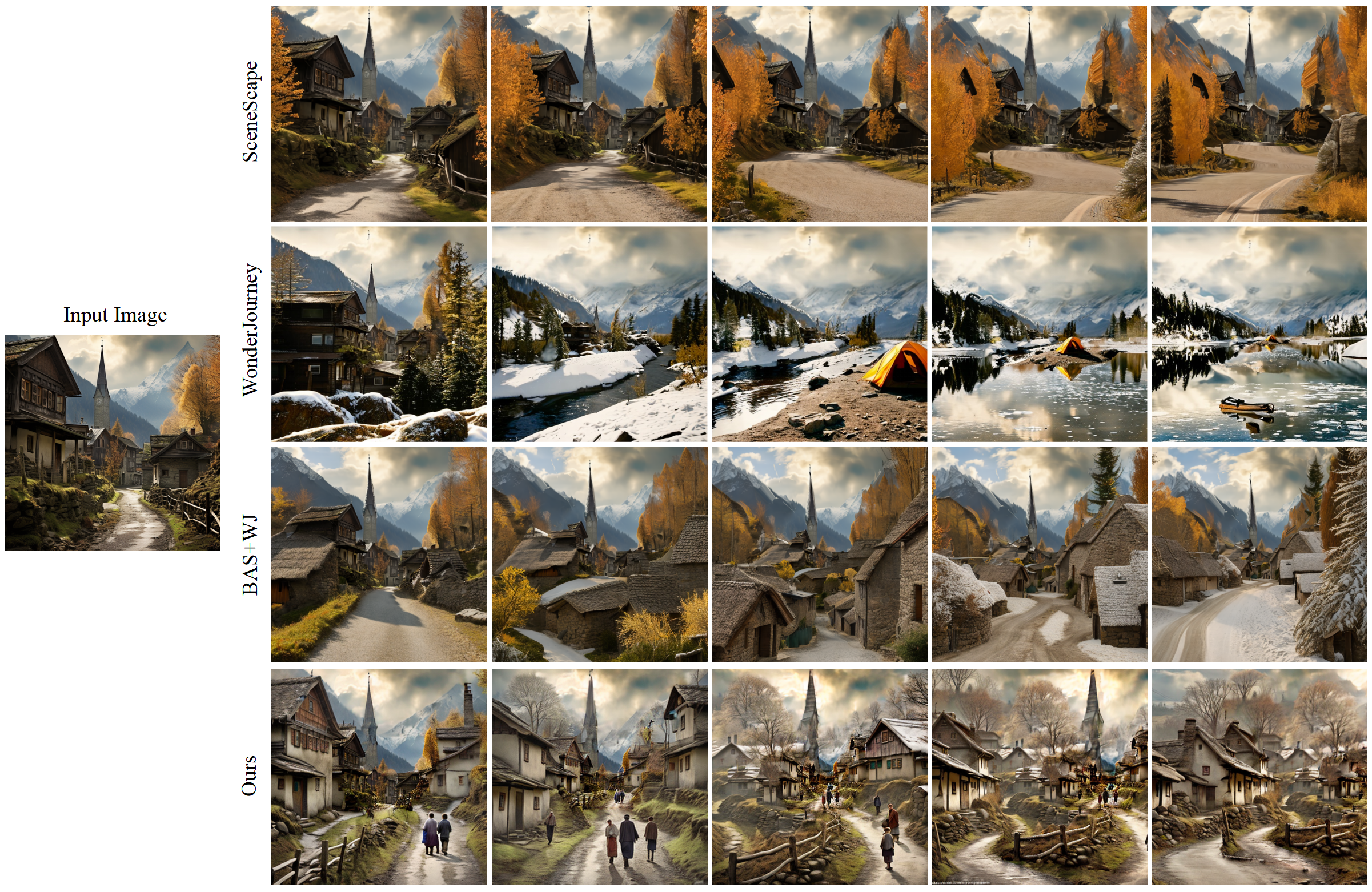}
    \caption{\textbf{Comparison with WonderJourney~\cite{yu2024wonderjourney} and SceneScape~\cite{fridman2024scenescape}.} We simply combine multi-concept customization method Break-A-Scene and 3D views generation framework WonderJourney as the baseline BAS+WJ for a comprehensive comparison.}
    \label{fig:comparison}
    \vspace{-2mm}
\end{figure*}
To evaluate the performance of our approach, we compare with previous methods in two tasks: Single Image Customization and 3D Views Generation. For the Single Image Customization task, we leverage the constructed SceneConceptGraph with optimized text-to-image model and choose four preeminent customization approaches to compare the quality and fidelity of the generated image: InstantBooth~\cite{shi2024instantbooth}, IP-Adapter~\cite{ye2023ip}, Custom Diffusion~\cite{kumari2023multi}and Break-A-Scene~\cite{avrahami2023break}. For the 3D Views Generation task, we maintain both construction and refinement processes and select two state-of-the-art methods to compare the performance of generated view sequences: SceneScape~\cite{fridman2024scenescape} and WonderJourney~\cite{yu2024wonderjourney}. We collect a dataset of 30 scenes, including nature, village, city, indoor room, or fantasy scenes, \etc and based on the scene dataset, we conduct extensive qualitative and quantitative comparison in the following parts.

\subsection{Qualitative Comparison}
\myparagraph{Single Image Customization.} We show qualitative comparisons with several mainstream customization strategies in Figure \ref{fig:comparisonbas}. We can observe that previous works fail to preserve the scene identity or be almost identical to the original scene. By contrast, our model is able to generate scene views with great fidelity and novel combination ways of scene concepts, which achieves high-quality scene-level customization and provides the scene-specific prior for the outpainting model in the following scene generation stage.

\myparagraph{3D Views Generation.}
We present qualitative examples of the generated 3D scene views compared to previous state-of-the-art methods in Figure \ref{fig:comparison}. It shows that with the generation going on, SceneScape~\cite{fridman2024scenescape} fails to create content-rich views, and WonderJourney~\cite{yu2024wonderjourney} suffers heavily from the semantic drift issue. Even with multi-concept customization on the first view, the simple combination baseline BAS+WJ still encounters problems including monotonous layouts, shifted overall style, and disorder in geometry due to the absence of the multi-level concepts and relations customization and test-time refinement. In contrast, our model generates more consistent and visually diverse scene views. 

\begin{figure*}[t]
    \centering
    \includegraphics[width=1.0\linewidth]{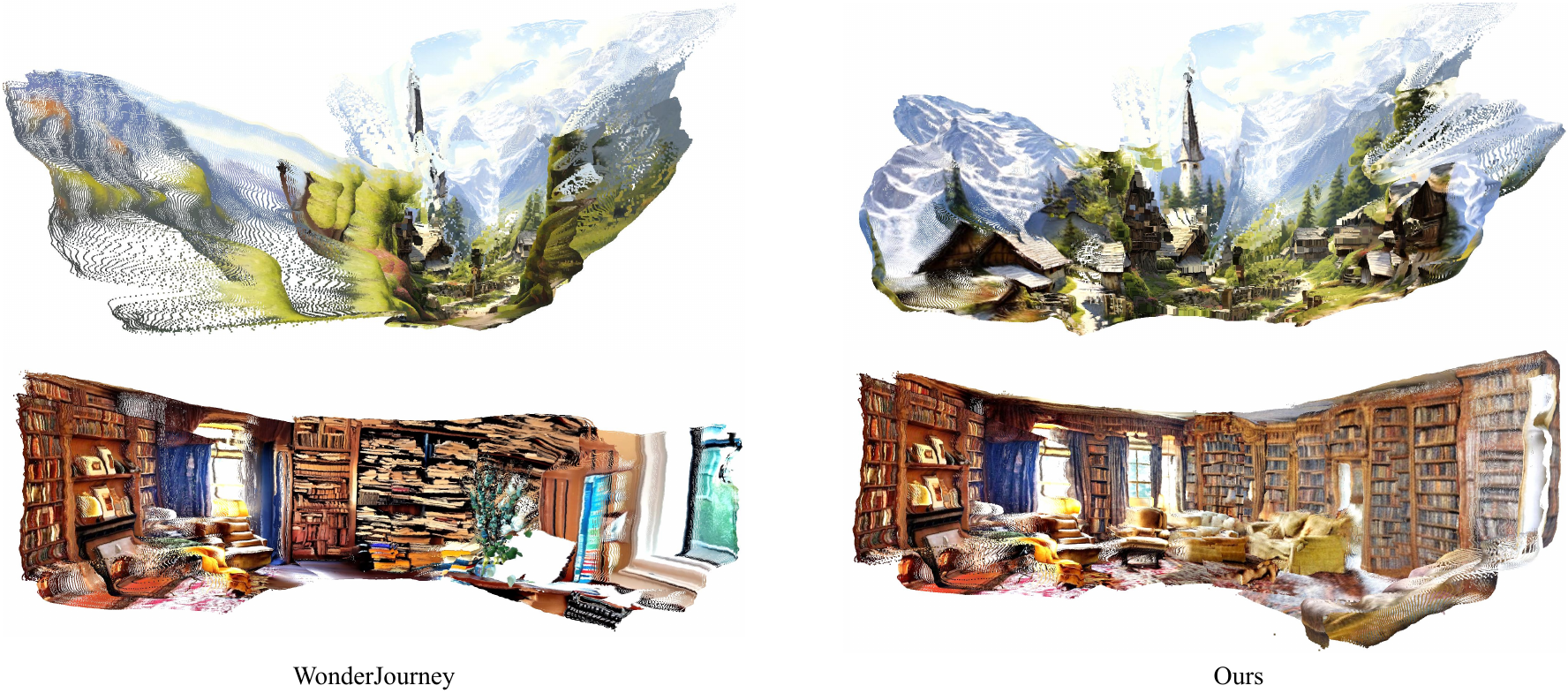}
    \caption{\textbf{3D representation comparison with WonderJourney~\cite{yu2024wonderjourney}. } We use the same first view and fixed camera path for evaluation. The generated 3D view sequences are shown in the supplementary material.}
    \label{fig:3d}
\end{figure*}

\begin{figure}[t]
    \centering
    \includegraphics[width=1.0\linewidth]{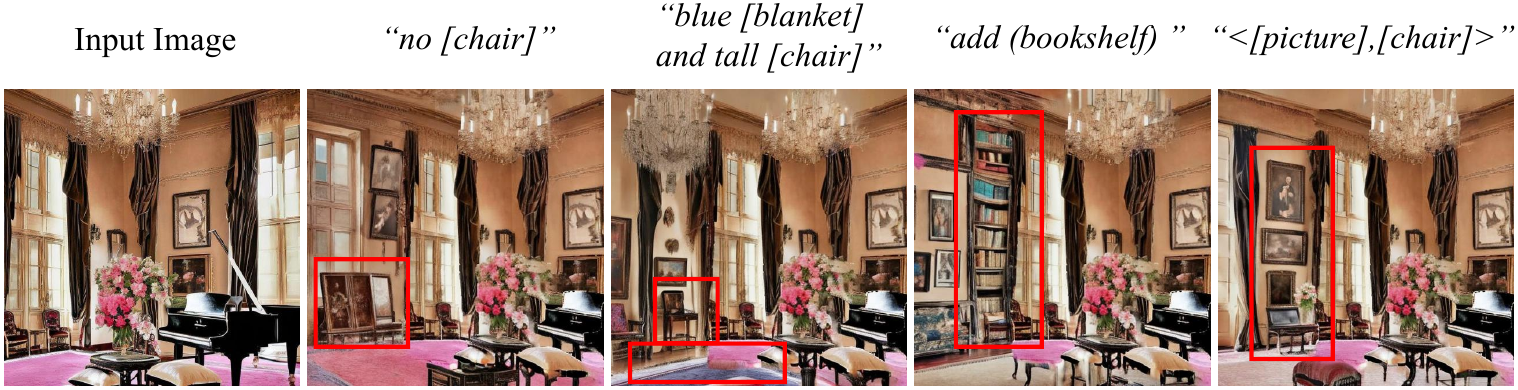}

    \caption{\textbf{Diversity and instruction fidelity of our  approach.} We use square brackets to denote defined concepts, parentheses to denote new objects, and angle brackets to indicate the relations between concepts.}
    \label{fig:dynamic}

\end{figure}

Furthermore, we present the intermediate 3D representation generated by WonderJourney and ours in Figure \ref{fig:3d}, which demonstrates that even though we adopt a similar unprojecting and rendering pipeline as WonderJourney, the inconsistency in objects and surroundings would cause the distortion in 3D geometric structure. Moreover, we provide a diverse example in Figure \ref{fig:dynamic} to illustrate how to generate the desired 3D views based on user instructions. More results of our 3D view sequences, constructed 3D scenes, and synthesized 3D videos can be found in the supplementary document and demo video.

\subsection{Quantitative Comparison}
\myparagraph{Single Image Customization.}
To quantitatively assess our method and the baseline, we focus on scene fidelity evaluation, which measures the preservation of initial scene details in generated images. We adopt CLIP-I~\cite{radford2021learning} and DINO~\cite{caron2021emerging} as evaluation metrics. These two metrics are the average pairwise cosine similarity of CLIP or ViT-S/16 DINO embeddings between the initial image and the generated image. As shown in Table \ref{tab:fidelity}, our method achieves the best scores in both CLIP-I and DINO metrics, indicating its high fidelity and strong scene-level customization capacity.

\begin{table}[t]
    \centering
    \caption{\textbf{Quantitative comparison of scene fidelity based on DINO and CLIP-I metrics.} Our method achieves the best scores.}
    \begin{tabular}{l|cc}
        \toprule
        Method & DINO$\ \uparrow$ & CLIP-I$\ \uparrow$ \\
        \midrule
        InstantBooth &0.785&0.832 \\
        IP-Adapter &0.899&0.922 \\
        Custom Diffusion &0.835&0.888\\
        Break-A-Scene&0.877&0.896\\
        ScenePainter (Ours)&\textbf{0.931}&\textbf{0.952} \\
        \bottomrule
    \end{tabular}
    \vspace{-2mm}
    \label{tab:fidelity}
\end{table}

\begin{table}[t]
    \centering
    \caption{\textbf{Quantitative comparison of human preference based on visual quality, diversity and consistency criteria.}}
    \begin{tabular}{l|ccc}
        \toprule
         & Qual. & Div. & Con.\\
        \midrule
        Ours over WonderJourney &89.3\%&83.4\%&92.6\%\\
        Ours over SceneScape &92.1\%&97.6\%&93.8\%\\
        \bottomrule
    \end{tabular}
    \vspace{-2mm}
    \label{tab:user}
\end{table}

\myparagraph{3D Views Generation.}
Since perpetual 3D scene generation is a new task without an existing dataset for quantitative evaluation, we conduct a user study on the subset of our collected dataset and focus on user preference evaluation. We generate scene views following each approach’s own setup and utilize visual quality, diversity and consistency as the evaluation metric. We show a side-by-side comparison of scene views generated by ours and SceneScape~\cite{fridman2024scenescape} or WonderJourney~\cite{yu2024wonderjourney} and then ask one binary choice question for users to decide. As shown in Table \ref{tab:user}, our model is strongly preferred over both baselines for all three metrics. WonderJourney generates wildly imaginative scenes, thus achieving acceptable diversity but notably low consistency caused by semantic deviation, whereas SceneScape generates more consistent but monotonous scenes due to the limited outpainting white space. Overall, our method achieves a well diversity-consistency balance and generates immersive 3D scene views.

\begin{table}[t]
    \centering
    \caption{\textbf{Quantitative comparison results of the ablation study on the loss function and the graph structure.}}
    \begin{tabular}{l|cc}
        \toprule
        Method & DINO$\ \uparrow$ & CLIP-I$\ \uparrow$ \\
        \midrule
        Ours w/o $\mathcal{L}_{\mathrm{rec}}$&0.523&0.568 \\
        Ours w/o $\mathcal{L}_{\mathrm{prior}}$&0.894&0.913 \\
        Ours w/o $\mathcal{L}_{\mathrm{attn}}$&0.723&0.768 \\
        Ours w/o $\mathcal{V}_1$&0.774&0.795 \\
        Ours w/o $\mathcal{V}_3$&0.857&0.886 \\
        Ours w/o $\mathcal{R}$&0.634&0.672 \\
        ScenePainter (Ours)&\textbf{0.924}&\textbf{0.949} \\
        \bottomrule
    \end{tabular}
    \vspace{-2mm}
    \label{tab:ablationtab}
\end{table}

\subsection{Ablation Study}
\begin{figure}[t]
    \centering
    \includegraphics[width=1.0\linewidth]{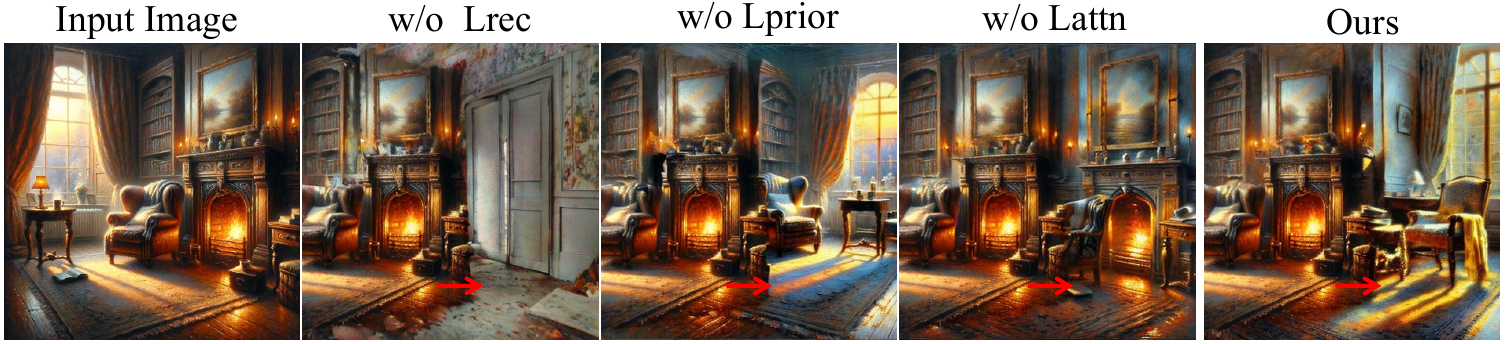}
    \caption{\textbf{Qualitative comparison results of the ablation study on the construction loss function.}}
    \label{fig:ablation}
    \vspace{-2mm}
\end{figure}

\begin{figure}[t]
    \centering
    \includegraphics[width=1.0\linewidth]{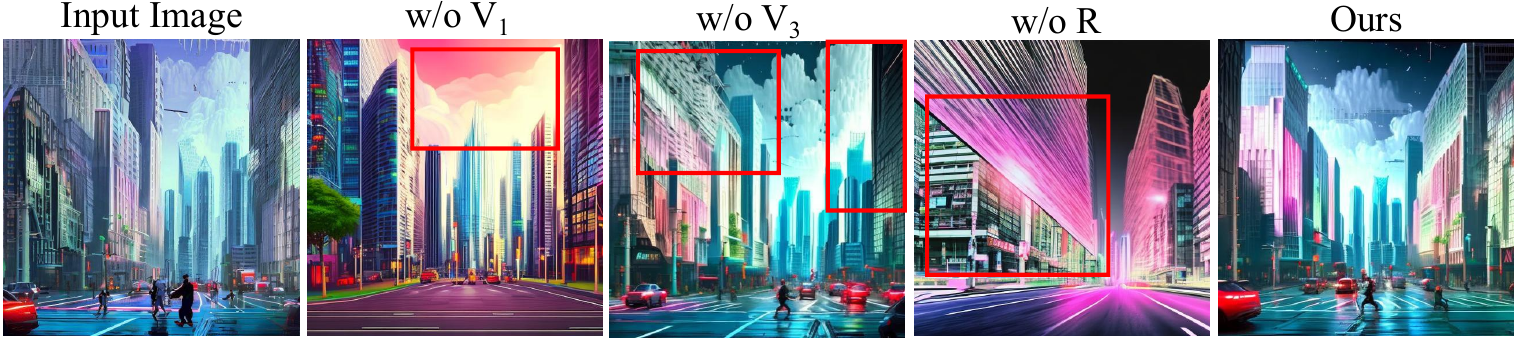}
    \caption{\textbf{Qualitative comparison results of the ablation study on the SceneConceptGraph structure.}}
    \label{fig:ablation1}
    \vspace{-2mm}
\end{figure}

We further conduct ablation studies to evaluate the validity of our method on the construction loss function Equation \ref{loss} and the structure of our proposed SceneConceptGraph. From Figure \ref{fig:ablation} and Table \ref{tab:ablationtab}, we observe that $\mathcal{L}_{rec}$ mainly ensures visual quality, $\mathcal{L}_{prior}$ inspires the generation of objects with different forms and $\mathcal{L}_{attn}$ prevents unreasonable object distributions and combinations. It can also be shown in Figure \ref{fig:ablation1} and Table \ref{tab:ablationtab} that removing the first-level concept the whole environment leads to the general style transfer, removing the third-level concept the individual object causes overfitting to the initial scene image layout and construction without relations causes distorted layouts and chaotic geometric structures. 

\subsection{Implementation Details}

In the concept relation construction and refinement stage, we leverage the Stable Diffusion~\cite{rombach2022high} model as the base text-to-image and outpainting model and SAM~\cite{kirillov2023segment} for mask segmentation. We customize the base model with the concept-relation pair prompt in two phases. In the first phase, we follow Textual Inversion and train text encoder with 400 steps, $1\times 10^{-6}$ learning rate. In the second phase, we follow DreamBooth and train the whole diffusion model with 400 steps, $1\times 10^{-4}$ learning rate. Then during the refinement process, we train the UNet module simply with 50 steps, 1e-4 learning rate for test-time efficiency. The training is done using a single NVIDIA A6000 GPU with about 5 minutes for construction and 25 seconds for refinement. For the entire 3D scene generation pipeline, we follow the similar unprojecting and rendering process proposed by WonderJourney with our customized model for outpainting and we adopt Blended Latent Diffusion~\cite{avrahami2023blended} for converting the text-to-image generation model to the outpainting model for scene-specific prior transfer.

\section{Conclusion}

In this paper, we propose ScenePainter, a new framework designed for semantically consistent and visually diverse 3D scene generation. To be specific, we construct and dynamically maintain the hierarchical graph focusing on the relations and concepts of the scene representation. Then we take it as dedicated textual embeddings to align and refine the outpainter's scene-specific prior with initial scene presentation for consistent expansion. Extensive experiments show that compared to previous methods, our framework effectively eliminates the semantic drift problem and produces a series of more consistent and vivid 3D scene views.

\section{Acknowledgement}

This work was supported by the National Natural Science Foundation of China under Grant 62206147.

{
    \small
    \bibliographystyle{ieeenat_fullname}
    \bibliography{main}

\begin{thebibliography}{39}
\providecommand{\natexlab}[1]{#1}
\providecommand{\url}[1]{\texttt{#1}}
\expandafter\ifx\csname urlstyle\endcsname\relax
  \providecommand{\doi}[1]{doi: #1}\else
  \providecommand{\doi}{doi: \begingroup \urlstyle{rm}\Url}\fi

\bibitem[Avrahami et~al.(2023{\natexlab{a}})Avrahami, Aberman, Fried, Cohen-Or, and Lischinski]{avrahami2023break}
Omri Avrahami, Kfir Aberman, Ohad Fried, Daniel Cohen-Or, and Dani Lischinski.
\newblock Break-a-scene: Extracting multiple concepts from a single image.
\newblock In \emph{SIGGRAPH Asia 2023 Conference Papers}, pages 1--12, 2023{\natexlab{a}}.

\bibitem[Avrahami et~al.(2023{\natexlab{b}})Avrahami, Fried, and Lischinski]{avrahami2023blended}
Omri Avrahami, Ohad Fried, and Dani Lischinski.
\newblock Blended latent diffusion.
\newblock \emph{ACM transactions on graphics (TOG)}, 42\penalty0 (4):\penalty0 1--11, 2023{\natexlab{b}}.

\bibitem[Blattmann et~al.(2023)Blattmann, Dockhorn, Kulal, Mendelevitch, Kilian, Lorenz, Levi, English, Voleti, Letts, et~al.]{blattmann2023stable}
Andreas Blattmann, Tim Dockhorn, Sumith Kulal, Daniel Mendelevitch, Maciej Kilian, Dominik Lorenz, Yam Levi, Zion English, Vikram Voleti, Adam Letts, et~al.
\newblock Stable video diffusion: Scaling latent video diffusion models to large datasets.
\newblock \emph{arXiv preprint arXiv:2311.15127}, 2023.

\bibitem[Brooks et~al.(2024)Brooks, Peebles, Holmes, DePue, Guo, Jing, Schnurr, Taylor, Luhman, Luhman, et~al.]{brooks2024video}
Tim Brooks, Bill Peebles, Connor Holmes, Will DePue, Yufei Guo, Li Jing, David Schnurr, Joe Taylor, Troy Luhman, Eric Luhman, et~al.
\newblock Video generation models as world simulators, 2024.

\bibitem[Cai et~al.(2023)Cai, Chan, Peng, Shahbazi, Obukhov, Van~Gool, and Wetzstein]{cai2023diffdreamer}
Shengqu Cai, Eric~Ryan Chan, Songyou Peng, Mohamad Shahbazi, Anton Obukhov, Luc Van~Gool, and Gordon Wetzstein.
\newblock Diffdreamer: Towards consistent unsupervised single-view scene extrapolation with conditional diffusion models.
\newblock In \emph{Proceedings of the IEEE/CVF International Conference on Computer Vision}, pages 2139--2150, 2023.

\bibitem[Caron et~al.(2021)Caron, Touvron, Misra, J{\'e}gou, Mairal, Bojanowski, and Joulin]{caron2021emerging}
Mathilde Caron, Hugo Touvron, Ishan Misra, Herv{\'e} J{\'e}gou, Julien Mairal, Piotr Bojanowski, and Armand Joulin.
\newblock Emerging properties in self-supervised vision transformers.
\newblock In \emph{Proceedings of the IEEE/CVF international conference on computer vision}, pages 9650--9660, 2021.

\bibitem[Chai et~al.(2023)Chai, Tucker, Li, Isola, and Snavely]{chai2023persistent}
Lucy Chai, Richard Tucker, Zhengqi Li, Phillip Isola, and Noah Snavely.
\newblock Persistent nature: A generative model of unbounded 3d worlds.
\newblock In \emph{Proceedings of the IEEE/CVF conference on computer vision and pattern recognition}, pages 20863--20874, 2023.

\bibitem[Chen et~al.(2023)Chen, Xia, He, Zhang, Cun, Yang, Xing, Liu, Chen, Wang, et~al.]{chen2023videocrafter1}
Haoxin Chen, Menghan Xia, Yingqing He, Yong Zhang, Xiaodong Cun, Shaoshu Yang, Jinbo Xing, Yaofang Liu, Qifeng Chen, Xintao Wang, et~al.
\newblock Videocrafter1: Open diffusion models for high-quality video generation.
\newblock \emph{arXiv preprint arXiv:2310.19512}, 2023.

\bibitem[Chung et~al.(2023)Chung, Lee, Nam, Lee, and Lee]{chung2023luciddreamer}
Jaeyoung Chung, Suyoung Lee, Hyeongjin Nam, Jaerin Lee, and Kyoung~Mu Lee.
\newblock Luciddreamer: Domain-free generation of 3d gaussian splatting scenes.
\newblock \emph{arXiv preprint arXiv:2311.13384}, 2023.

\bibitem[Fridman et~al.(2024)Fridman, Abecasis, Kasten, and Dekel]{fridman2024scenescape}
Rafail Fridman, Amit Abecasis, Yoni Kasten, and Tali Dekel.
\newblock Scenescape: Text-driven consistent scene generation.
\newblock \emph{Advances in Neural Information Processing Systems}, 36, 2024.

\bibitem[Gal et~al.(2022)Gal, Alaluf, Atzmon, Patashnik, Bermano, Chechik, and Cohen-Or]{gal2022image}
Rinon Gal, Yuval Alaluf, Yuval Atzmon, Or Patashnik, Amit~H Bermano, Gal Chechik, and Daniel Cohen-Or.
\newblock An image is worth one word: Personalizing text-to-image generation using textual inversion.
\newblock \emph{arXiv preprint arXiv:2208.01618}, 2022.

\bibitem[Han et~al.(2023)Han, Li, Zhang, Milanfar, Metaxas, and Yang]{han2023svdiff}
Ligong Han, Yinxiao Li, Han Zhang, Peyman Milanfar, Dimitris Metaxas, and Feng Yang.
\newblock Svdiff: Compact parameter space for diffusion fine-tuning.
\newblock In \emph{Proceedings of the IEEE/CVF International Conference on Computer Vision}, pages 7323--7334, 2023.

\bibitem[Ho et~al.(2020)Ho, Jain, and Abbeel]{ho2020denoising}
Jonathan Ho, Ajay Jain, and Pieter Abbeel.
\newblock Denoising diffusion probabilistic models.
\newblock \emph{Advances in neural information processing systems}, 33:\penalty0 6840--6851, 2020.

\bibitem[Hu et~al.(2021)Hu, Shen, Wallis, Allen-Zhu, Li, Wang, Wang, and Chen]{hu2021lora}
Edward~J Hu, Yelong Shen, Phillip Wallis, Zeyuan Allen-Zhu, Yuanzhi Li, Shean Wang, Lu Wang, and Weizhu Chen.
\newblock Lora: Low-rank adaptation of large language models.
\newblock \emph{arXiv preprint arXiv:2106.09685}, 2021.

\bibitem[Jain et~al.(2024)Jain, Nasery, Vineet, and Behl]{jain2024peekaboo}
Yash Jain, Anshul Nasery, Vibhav Vineet, and Harkirat Behl.
\newblock Peekaboo: Interactive video generation via masked-diffusion.
\newblock In \emph{Proceedings of the IEEE/CVF Conference on Computer Vision and Pattern Recognition}, pages 8079--8088, 2024.

\bibitem[Kaneva et~al.(2010)Kaneva, Sivic, Torralba, Avidan, and Freeman]{kaneva2010infinite}
Biliana Kaneva, Josef Sivic, Antonio Torralba, Shai Avidan, and William~T Freeman.
\newblock Infinite images: Creating and exploring a large photorealistic virtual space.
\newblock \emph{Proceedings of the IEEE}, 98\penalty0 (8):\penalty0 1391--1407, 2010.

\bibitem[Kerbl et~al.(2023)Kerbl, Kopanas, Leimk{\"u}hler, and Drettakis]{kerbl20233d}
Bernhard Kerbl, Georgios Kopanas, Thomas Leimk{\"u}hler, and George Drettakis.
\newblock 3d gaussian splatting for real-time radiance field rendering.
\newblock \emph{ACM Trans. Graph.}, 42\penalty0 (4):\penalty0 139--1, 2023.

\bibitem[Kirillov et~al.(2023)Kirillov, Mintun, Ravi, Mao, Rolland, Gustafson, Xiao, Whitehead, Berg, Lo, et~al.]{kirillov2023segment}
Alexander Kirillov, Eric Mintun, Nikhila Ravi, Hanzi Mao, Chloe Rolland, Laura Gustafson, Tete Xiao, Spencer Whitehead, Alexander~C Berg, Wan-Yen Lo, et~al.
\newblock Segment anything.
\newblock In \emph{Proceedings of the IEEE/CVF international conference on computer vision}, pages 4015--4026, 2023.

\bibitem[Kumari et~al.(2023)Kumari, Zhang, Zhang, Shechtman, and Zhu]{kumari2023multi}
Nupur Kumari, Bingliang Zhang, Richard Zhang, Eli Shechtman, and Jun-Yan Zhu.
\newblock Multi-concept customization of text-to-image diffusion.
\newblock In \emph{Proceedings of the IEEE/CVF Conference on Computer Vision and Pattern Recognition}, pages 1931--1941, 2023.

\bibitem[Li et~al.(2022)Li, Wang, Snavely, and Kanazawa]{li2022infinitenature}
Zhengqi Li, Qianqian Wang, Noah Snavely, and Angjoo Kanazawa.
\newblock Infinitenature-zero: Learning perpetual view generation of natural scenes from single images.
\newblock In \emph{European Conference on Computer Vision}, pages 515--534. Springer, 2022.

\bibitem[Liu et~al.(2021)Liu, Tucker, Jampani, Makadia, Snavely, and Kanazawa]{liu2021infinite}
Andrew Liu, Richard Tucker, Varun Jampani, Ameesh Makadia, Noah Snavely, and Angjoo Kanazawa.
\newblock Infinite nature: Perpetual view generation of natural scenes from a single image.
\newblock In \emph{Proceedings of the IEEE/CVF International Conference on Computer Vision}, pages 14458--14467, 2021.

\bibitem[Podell et~al.(2023)Podell, English, Lacey, Blattmann, Dockhorn, M{\"u}ller, Penna, and Rombach]{podell2023sdxl}
Dustin Podell, Zion English, Kyle Lacey, Andreas Blattmann, Tim Dockhorn, Jonas M{\"u}ller, Joe Penna, and Robin Rombach.
\newblock Sdxl: Improving latent diffusion models for high-resolution image synthesis.
\newblock \emph{arXiv preprint arXiv:2307.01952}, 2023.

\bibitem[Radford et~al.(2021)Radford, Kim, Hallacy, Ramesh, Goh, Agarwal, Sastry, Askell, Mishkin, Clark, et~al.]{radford2021learning}
Alec Radford, Jong~Wook Kim, Chris Hallacy, Aditya Ramesh, Gabriel Goh, Sandhini Agarwal, Girish Sastry, Amanda Askell, Pamela Mishkin, Jack Clark, et~al.
\newblock Learning transferable visual models from natural language supervision.
\newblock In \emph{International conference on machine learning}, pages 8748--8763. PMLR, 2021.

\bibitem[Rombach et~al.(2022)Rombach, Blattmann, Lorenz, Esser, and Ommer]{rombach2022high}
Robin Rombach, Andreas Blattmann, Dominik Lorenz, Patrick Esser, and Bj{\"o}rn Ommer.
\newblock High-resolution image synthesis with latent diffusion models.
\newblock In \emph{Proceedings of the IEEE/CVF conference on computer vision and pattern recognition}, pages 10684--10695, 2022.

\bibitem[Ruiz et~al.(2023)Ruiz, Li, Jampani, Pritch, Rubinstein, and Aberman]{ruiz2023dreambooth}
Nataniel Ruiz, Yuanzhen Li, Varun Jampani, Yael Pritch, Michael Rubinstein, and Kfir Aberman.
\newblock Dreambooth: Fine tuning text-to-image diffusion models for subject-driven generation.
\newblock In \emph{Proceedings of the IEEE/CVF conference on computer vision and pattern recognition}, pages 22500--22510, 2023.

\bibitem[Ryu()]{ryulow}
Simo Ryu.
\newblock Low-rank adaptation for fast text-to-image diffusion fine-tuning. 2022.
\newblock \emph{URL https://github. com/cloneofsimo/lora}.

\bibitem[Shi et~al.(2024)Shi, Xiong, Lin, and Jung]{shi2024instantbooth}
Jing Shi, Wei Xiong, Zhe Lin, and Hyun~Joon Jung.
\newblock Instantbooth: Personalized text-to-image generation without test-time finetuning.
\newblock In \emph{CVPR}, pages 8543--8552, 2024.

\bibitem[Voynov et~al.(2023)Voynov, Chu, Cohen-Or, and Aberman]{voynov2023p+}
Andrey Voynov, Qinghao Chu, Daniel Cohen-Or, and Kfir Aberman.
\newblock p+: Extended textual conditioning in text-to-image generation.
\newblock \emph{arXiv preprint arXiv:2303.09522}, 2023.

\bibitem[Wang et~al.(2024{\natexlab{a}})Wang, Xiang, Fan, and Xue]{wang2024customizing}
Hai Wang, Xiaoyu Xiang, Yuchen Fan, and Jing-Hao Xue.
\newblock Customizing 360-degree panoramas through text-to-image diffusion models.
\newblock In \emph{Proceedings of the IEEE/CVF Winter Conference on Applications of Computer Vision}, pages 4933--4943, 2024{\natexlab{a}}.

\bibitem[Wang et~al.(2024{\natexlab{b}})Wang, Yuan, Wang, Li, Chen, Xia, Luo, and Shan]{wang2024motionctrl}
Zhouxia Wang, Ziyang Yuan, Xintao Wang, Yaowei Li, Tianshui Chen, Menghan Xia, Ping Luo, and Ying Shan.
\newblock Motionctrl: A unified and flexible motion controller for video generation.
\newblock In \emph{ACM SIGGRAPH 2024 Conference Papers}, pages 1--11, 2024{\natexlab{b}}.

\bibitem[Wei et~al.(2023)Wei, Zhang, Ji, Bai, Zhang, and Zuo]{wei2023elite}
Yuxiang Wei, Yabo Zhang, Zhilong Ji, Jinfeng Bai, Lei Zhang, and Wangmeng Zuo.
\newblock Elite: Encoding visual concepts into textual embeddings for customized text-to-image generation.
\newblock In \emph{Proceedings of the IEEE/CVF International Conference on Computer Vision}, pages 15943--15953, 2023.

\bibitem[Yang et~al.(2024)Yang, Tan, Zhang, Wu, Li, Wetzstein, Liu, and Lin]{yang2024layerpano3d}
Shuai Yang, Jing Tan, Mengchen Zhang, Tong Wu, Yixuan Li, Gordon Wetzstein, Ziwei Liu, and Dahua Lin.
\newblock Layerpano3d: Layered 3d panorama for hyper-immersive scene generation.
\newblock \emph{arXiv preprint arXiv:2408.13252}, 2024.

\bibitem[Yatim et~al.(2024)Yatim, Fridman, Bar-Tal, Kasten, and Dekel]{yatim2024space}
Danah Yatim, Rafail Fridman, Omer Bar-Tal, Yoni Kasten, and Tali Dekel.
\newblock Space-time diffusion features for zero-shot text-driven motion transfer.
\newblock In \emph{Proceedings of the IEEE/CVF Conference on Computer Vision and Pattern Recognition}, pages 8466--8476, 2024.

\bibitem[Ye et~al.(2023)Ye, Zhang, Liu, Han, and Yang]{ye2023ip}
Hu Ye, Jun Zhang, Sibo Liu, Xiao Han, and Wei Yang.
\newblock Ip-adapter: Text compatible image prompt adapter for text-to-image diffusion models.
\newblock \emph{arXiv preprint arXiv:2308.06721}, 2023.

\bibitem[Yu et~al.(2024{\natexlab{a}})Yu, Duan, Herrmann, Freeman, and Wu]{yu2024wonderworld}
Hong-Xing Yu, Haoyi Duan, Charles Herrmann, William~T Freeman, and Jiajun Wu.
\newblock Wonderworld: Interactive 3d scene generation from a single image.
\newblock \emph{arXiv preprint arXiv:2406.09394}, 2024{\natexlab{a}}.

\bibitem[Yu et~al.(2024{\natexlab{b}})Yu, Duan, Hur, Sargent, Rubinstein, Freeman, Cole, Sun, Snavely, Wu, et~al.]{yu2024wonderjourney}
Hong-Xing Yu, Haoyi Duan, Junhwa Hur, Kyle Sargent, Michael Rubinstein, William~T Freeman, Forrester Cole, Deqing Sun, Noah Snavely, Jiajun Wu, et~al.
\newblock Wonderjourney: Going from anywhere to everywhere.
\newblock In \emph{Proceedings of the IEEE/CVF Conference on Computer Vision and Pattern Recognition}, pages 6658--6667, 2024{\natexlab{b}}.

\bibitem[Zhang et~al.(2024)Zhang, Wu, Gambardella, Huang, Phung, Ouyang, and Cai]{zhang2024taming}
Cheng Zhang, Qianyi Wu, Camilo~Cruz Gambardella, Xiaoshui Huang, Dinh Phung, Wanli Ouyang, and Jianfei Cai.
\newblock Taming stable diffusion for text to 360 panorama image generation.
\newblock In \emph{Proceedings of the IEEE/CVF Conference on Computer Vision and Pattern Recognition}, pages 6347--6357, 2024.

\bibitem[Zhou et~al.(2024)Zhou, Cheng, Yu, Tian, and Yuan]{zhou2024holodreamer}
Haiyang Zhou, Xinhua Cheng, Wangbo Yu, Yonghong Tian, and Li Yuan.
\newblock Holodreamer: Holistic 3d panoramic world generation from text descriptions.
\newblock \emph{arXiv preprint arXiv:2407.15187}, 2024.

\bibitem[Zhou et~al.(2025)Zhou, Fan, Xu, Chang, Chari, Bharadwaj, You, Wang, and Kadambi]{zhou2025dreamscene360}
Shijie Zhou, Zhiwen Fan, Dejia Xu, Haoran Chang, Pradyumna Chari, Tejas Bharadwaj, Suya You, Zhangyang Wang, and Achuta Kadambi.
\newblock Dreamscene360: Unconstrained text-to-3d scene generation with panoramic gaussian splatting.
\newblock In \emph{European Conference on Computer Vision}, pages 324--342. Springer, 2025.

\end{thebibliography}
}

\end{document}